\documentclass[sigconf]{acmart} %review
\makeatletter
\def\@ACM@checkaffil{% Only warnings
    \if@ACM@instpresent\else
    \ClassWarningNoLine{\@classname}{No institution present for an affiliation}%
    \fi
    \if@ACM@citypresent\else
    \ClassWarningNoLine{\@classname}{No city present for an affiliation}%
    \fi
    \if@ACM@countrypresent\else
        \ClassWarningNoLine{\@classname}{No country present for an affiliation}%
    \fi
}
\makeatother

\usepackage{algorithm}
\usepackage{algorithmicx}
\usepackage{algpseudocode}
\usepackage{multirow}
\usepackage{enumitem}
\usepackage{graphicx}
\usepackage{subfigure}
\usepackage{soul}

\usepackage{xspace}
\newcommand{\etal}{\emph{et al.}\xspace}
\newcommand{\eg}{\emph{e.g.,}\xspace}
\newcommand{\ie}{\emph{i.e.,}\xspace}
\newcommand{\etc}{\emph{etc.}\xspace}
\newcommand{\name}{PromptST\xspace}
\newcommand{\namewospace}{PromptST}
\newcommand{\stpmt}{spatio-temporal prompt\xspace}
\newcommand{\apmt}{tiny spatio-temporal prompt\xspace}
\newcommand{\stpmthead}{Spatio-Temporal Prompt\xspace}

\usepackage{color}

\AtBeginDocument{%
  \providecommand\BibTeX{{%
    \normalfont B\kern-0.5em{\scshape i\kern-0.25em b}\kern-0.8em\TeX}}}

\setcopyright{acmcopyright}
\copyrightyear{2018}
\acmYear{2018}
\acmDOI{XXXXXXX.XXXXXXX}

\acmPrice{15.00}
\acmISBN{978-1-4503-XXXX-X/18/06}

\author{Zijian Zhang}
\affiliation{%
  \institution{Jilin University}
  \institution{City University of Hong Kong}
   }
\email{zhangzj2114@mails.jlu.edu.cn}

\author{Xiangyu Zhao}
\affiliation{%
  \institution{City University of Hong Kong}
  }
\email{xianzhao@cityu.edu.hk}
\authornote{Corresponding authors: Xiangyu Zhao, Wanyu Wang and Hongwei Zhao}

\author{Qidong Liu}
\affiliation{%
  \institution{Xi'an Jiaotong University} 
  \institution{City University of Hong Kong}
}
\email{liuqidong@stu.xjtu.edu.cn}

\author{Chunxu Zhang}
\affiliation{%
  \institution{Jilin University}
  }
\email{cxzhang19@mails.jlu.edu.cn}

\author{Qian Ma}
\affiliation{%
  \institution{City University of Hong Kong}
  }
\email{qma233-c@my.cityu.edu.hk}

\author{Wanyu Wang}
\affiliation{%
  \institution{City University of Hong Kong}
  }
\email{wanyuwang4-c@my.cityu.edu.hk}

\authornotemark[1]
\author{Hongwei Zhao}
\affiliation{%
  \institution{Jilin University}
  }
\email{zhaohw@jlu.edu.cn}
\authornotemark[1]

\author{Yiqi Wang}
\affiliation{%
  \institution{Michigan State University}
  }
\email{wangy206@msu.edu} % 0000-0001-9594-1919

\author{Zitao Liu}
\affiliation{%
  \institution{
  Guangdong Institute of Smart Education, 
  Jinan University}
  }
\email{liuzitao@jnu.edu.cn} % 0000-0001-9594-1919

\begin{document}
\copyrightyear{2023}
\acmYear{2023}
\setcopyright{acmlicensed}\acmConference[CIKM '23]{Proceedings of the 32nd
ACM International Conference on Information and Knowledge
Management}{October 21--25, 2023}{Birmingham, United Kingdom}
\acmBooktitle{Proceedings of the 32nd ACM International Conference on
Information and Knowledge Management (CIKM '23), October 21--25, 2023,
Birmingham, United Kingdom}
\acmPrice{15.00}
\acmDOI{10.1145/3583780.3615016}
\acmISBN{979-8-4007-0124-5/23/10}
\renewcommand{\shortauthors}{Zijian Zhang, et al.}

\title{
% Towards Unified Smart City Modeling: Prompt Tuned Spatio-Temporal Multivariate Prediction\\
% Towards Unified Smart City Modeling: Prompt-Enhanced Spatio-Temporal Prediction\\
PromptST: Prompt-Enhanced Spatio-Temporal Multi-Attribute Prediction\\
% Towards Unified Spatio-Temporal Prediction: A Prompt Perspective
}

\begin{abstract}
In the era of information explosion, spatio-temporal data mining serves as a critical part of urban management. Considering the various fields demanding attention, \eg traffic state, human activity, and social event, predicting multiple spatio-temporal attributes simultaneously can alleviate regulatory pressure and foster smart city construction. However, current research can not handle the spatio-temporal multi-attribute prediction well due to the complex relationships between diverse attributes. The key challenge lies in how to address the common spatio-temporal patterns while tackling their distinctions. In this paper, we propose an effective solution for spatio-temporal multi-attribute prediction, \name. We devise a spatio-temporal transformer and a parameter-sharing training scheme to address the common knowledge among different spatio-temporal attributes. Then, we elaborate a spatio-temporal prompt tuning strategy to fit the specific attributes in a lightweight manner. Through the pretrain and prompt tuning phases, our \name is able to enhance the specific spatio-temoral characteristic capture by prompting the backbone model to fit the specific target attribute while maintaining the learned common knowledge. Extensive experiments on real-world datasets verify that our \name attains state-of-the-art performance. Furthermore, we also prove \name owns good transferability on unseen spatio-temporal attributes, which brings promising application potential in urban computing. The implementation code is available to ease reproducibility\footnote{\url{https://github.com/Zhang-Zijian/PromptST}}.
\end{abstract}

% <concept_id>10002951.10003227.10003236</concept_id>
% <concept_desc>Information systems~Spatial-temporal systems</concept_desc>
% <concept_id>10002951.10003260.10003277.10003281</concept_id>
% <concept_desc>Information systems~Traffic analysis</concept_desc>
% <concept_id>10010147.10010257.10010293.10010294</concept_id>
% <concept_desc>Computing methodologies~Neural networks</concept_desc>



\maketitle
\section{Introduction}

% <\textcolor{red}{The 1st paragraph: the background of intelligent transportation systems, then why we do the multi-variate traffic prediction.}>

Fast urbanization has brought unprecedented convenience to our lives, and also facilitated explosively growth in transportation, security incidents, social events \etc 
However, it pressures the authority to keep a watchful eye on all these issues \cite{smart_nature}, which requires burdensome expert efforts and economic costs.
Thanks to the prosperous development of Spatio-Temporal Data Mining (STDM), it has become a reality to analyze urban big data and predict future trends \cite{smart_stdm}, \eg intelligent transportation systems \cite{strn, 10.1145/3580305.3599528, wang2021traffic}, weather and air quality prediction \cite{han2021joint,  zhao2017incorporating}, and crime prediction \cite{aaai22, zhao2017exploring, zhao2018crime, zhao2017modeling, zhao2016exploring}.

\begin{figure}[!t]
\includegraphics[width=1\linewidth]{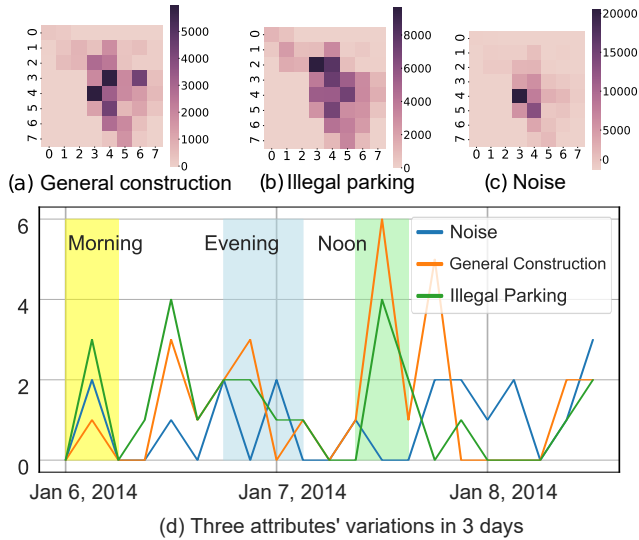}
% {\subfigure{\includegraphics[width=0.32\linewidth]{Figure/intro8.pdf}}}
% {\subfigure{\includegraphics[width=0.32\linewidth]{Figure/intro9.pdf}}}
% {\subfigure{\includegraphics[width=0.32\linewidth]{Figure/intro7.pdf}}}

% {\subfigure{\includegraphics[width=1\linewidth]{Figure/time_intro_d.pdf}}}
\vspace*{-4mm}
    \caption{Illustrations of three spatio-temporal attributes in Complaint dataset. (a), (b), and (c) indicate the amounts of the three attributes in $8\times 8$-sized city, respectively. 
    The darker the grid represents the more complaints of corresponding types. 
    (d) depicts the temporal variations of the three attributes.
    }
    \label{fig:intro}
% \vspace*{-3mm}
\end{figure}

The miscellaneous tasks required to be handled in city management emerge in complex spatio-temporal patterns.
Considering the function of different regions, \textit{human activities in different regions proceed in similar or different ways.}
Take regional complaint records from NYC 311\footnote{\url{https://portal.311.nyc.gov/}} as an example, Figure \ref{fig:intro} (a), (b), and (c) illustrate the complaints for `general construction', `illegal parking', and `noise', respectively.
Figure \ref{fig:intro} (a) and (b) share similar spatial patterns because building and parking problems generally arise in the city. 
In contrast, (c) owns a prominent high volume in grid $(4, 3)$, which can be ascribed to the high human mobility in the downtown area. 
Besides, we also find that \textit{different attributes change in similar or different trends over time.}
Figure \ref{fig:intro} (d) shows the three attributes' variations in three days. 
The three types of complaints all occur in the morning, but complaints of noise are reported more in the evening and the other two peak at noon. 
These multiple spatio-temporal attributes are intertwined, 
and analyzing and utilizing their correlations is of great significance for spatio-temporal multi-attribute prediction and unifying smart city management.
% \lqd{These two statements are challenges or findings?}

Currently, {existing} STDM works can not handle the spatio-temporal multi-attribute prediction well.
\textit{Spatio-temporal prediction methods} generally focus on higher performance on a single attribute \cite{yu2017spatio,zheng2020gman}, which integrates expert knowledge-based inductive bias \cite{wang2021traffic, zhang2020taxi} and devises highly-specific architectures for single tasks. 
For example, STGCN~\cite{yu2017spatio} utilizes the ChebNet graph convolution to predict traffic flows. GMAN~\cite{zheng2020gman} proposes a fabricated model equipped with an attention mechanism to forecast traffic volume or traffic speed.
This line of work ignores the relationship between multiple spatio-temporal attributes and fails to address multi-attribute prediction.
On the other hand, \textit{Spatio-temporal multi-task learning methods} pay attention to solving two specific tasks with similar spatio-temporal patterns, which is hard to be extended to numerous tasks due to the highly specified architecture.
For instance, MDL \cite{zhang2019flow} and pmlLSTM \cite{zhang2020taxi} are the representative attempts predicting two traffic flows simultaneously, where the convolutional neural networks and LSTM are incorporated, respectively.
GEML \cite{wang2019origin} and MT-ASTN \cite{wang2020multi} propose to handle traffic flow and on-demand flow, where the challenge is the feature fusion of different shapes.
Recently, AutoSTL \cite{autostl} shows promising capability on multiple spatio-temporal tasks. However, it also focuses on two tasks, leaving performance on numerous attributes unexplored.

%Recently, multi-task learning has proved its efficacy in related areas \cite{ma2018modeling, tang2020progressive, qin2020multitask}, which has an edge on learning correlations between multiple tasks. 
%MMoE~\cite{ma2018modeling} applies multiple expert networks to model homogeneous and heterogeneous information of different tasks and uses the gate mechanism to weight the contributions of each expert. 
%Based on MMoE, PLE~\cite{tang2020progressive} separates the shared and specific experts and captures common and task-specific dependency, respectively.
%Nevertheless, multi-task learning methods perform well with limited tasks and the model complexity increases with more tasks.
% However, the time and space cost of multi-task learning methods increase dramatically with the number of variates, which is not suitable for city management with numerous spatio-temporal attributes. 
% \lqd{MTL cost more?}

To shed light on the solution of spatio-temporal multi-attribute prediction, we recognize three properties that a spatio-temporal multi-attribute approach should own:
(i) \textbf{Generality}. To model the numerous spatio-temporal attributes, the model should well-capture the common characteristics among various attributes.
(ii) \textbf{Adaptivity}. Facing diverse spatio-temporal patterns, the model is supposed to be flexible to fit distinct characteristics of the specific attributes. 
(iii) \textbf{Transferability}. 
Considering the extensive application scenarios of spatio-temporal prediction, the method is supposed to be lightweight and easily extensible to new tasks. 
% \zxy{how about the temporal/computing Efficiency?}

% However, it is non-trivial to accomplish the goals, and the challenges lie in three aspects: 
% (i) To capture the common characteristic among multiple attributes, it demands high capacity and effective spatio-temporal modeling. 
% (ii) The model should address the specific attributes well while maintaining common knowledge. 
% (iii) The model is supposed to have advancing time and space efficiency, so that it could be deployed generally.

To accomplish the goals above, challenges in three aspects should be conquered. 
(i) Capturing the common characteristic among multiple attributes demands high capacity and effective spatio-temporal modeling. 
(ii) The model should address the specific attributes well while maintaining common knowledge. 
(iii) The model is supposed to have advancing time and space efficiency so as to be deployed generally.
Large pretrain models~\cite{brown2020language} seem to be a potential solution, which can enhance the performance on target attributes based on the common knowledge learned from large datasets.
However, while showing effectiveness over related tasks \cite{muppet, shao2022pre}, such pretrain and fine-tune pattern suffers from the large retraining cost and catastrophic forgetting \cite{ft_forget, ft_forget_origin}.

In this paper, we propose a pretrain and prompt tuning method, \name, to solve spatio-temporal multi-attribute prediction effectively.
We first put forward an effective spatio-temporal transformer and train it with multiple spatio-temporal attributes in a parameter-sharing manner. 
Thus, it can well-address the common knowledge of multiple spatio-temporal attributes.
% \zxy{Borrowing the idea from computer vision~\cite{vpt_pmt, ofa_pmt} and NLP~\cite{cdq_pmt, lpf_pmt}}, 
We devise novel \stpmt tokens and insert them into the feature sequence during target attribute tuning~\cite{cdq_pmt, lpf_pmt}. 
By fixing the main body of pretrained transformer and only tuning prompt tokens with head, our \name achieves a good balance between fitting the target spatio-temporal attribute and maintaining the common spatio-temporal knowledge.
{Common spatio-temporal knowledge can also enhance the modeling of new tasks with lightweight tuning, thus facilitating the analysis of urban attributes in more domains.}
Our method attains state-of-the-art performance on two real-world datasets {(with 19 and 4 different physical tasks respectively)} by only tuning 11\% of the backbone model parameters.
Besides, we explore a lightweight version of prompt tuning, \apmt, to further improve the time and space efficiency, which demands trivial (less than \textbf{1\%} of model parameters) trainable parameters. Our \name also shows good transferability to fit the unseen spatio-temporal attributes.
In a nutshell, our main contributions are as follows:

\begin{itemize}[leftmargin=*]
    % \item {We present a transformer-based backbone with multi-attribute training strategy to capture the common spatio-temporal patterns, and we elaborate a swift prompt tuning phase to fit specific attributes. To the best of the author's knowledge, our \name is the first pretrain-prompt tuning scheme to solve the spatio-temporal multi-attribute prediction;}
    
    \item {For the first time, we propose a pretrain-prompt tuning scheme, \name, to solve the spatio-temporal multi-attribute prediction. It can well fit the specific spatio-temporal pattern of multiple attributes with the enhancement of learned common knowledge;}
    % To the best of the author's knowledge, it is the first pretrain-prompt tuning scheme to solve the spatio-temporal prediction with numerous attributes;}

    %  It can balance the learned common knowledge and spatio-temporal pattern of specific target attributes.
    % \item We present a novel transformer-based framework with a parameter-sharing training strategy to capture the common spatio-temporal patterns in multiple attributes.
    % \zzj{To the best of the authors' knowledge, this is the first transformer-based backbone to address the common knowledge among numerous spatio-temporal attributes;}

    \item {Our lightweight spatio-temporal prompt tokens save almost 89\% trainable parameters in the tuning phase, with its tiny version achieving a reduction of 99\%. This immensely motivates the wide deployment of urban attribute analyzers;}
    
    % \item \zzj{We propose a pretrain-prompt tuning scheme to solve the spatio-temporal multi-attribute prediction, named \name. Our lightweight spatio-temporal prompt tokens save almost 89\% trainable parameters in the tuning phase, with its tiny version achieving a reduction of 99\%. This immensely motivates the wide deployment of urban attribute analyzers;}
    
    \item Extensive experiments on two real-world datasets verify the state-of-the-art performance of \name against diverse advancing baselines. Furthermore, {\name also possesses good transferability, which solves unseen attributes with trivial training costs and shows promising potential for real-world applications.}
\end{itemize}

\begin{figure*}[!ht]
\centering
	%	\hspace*{-7mm}1.049
	\includegraphics[width=0.6\linewidth]{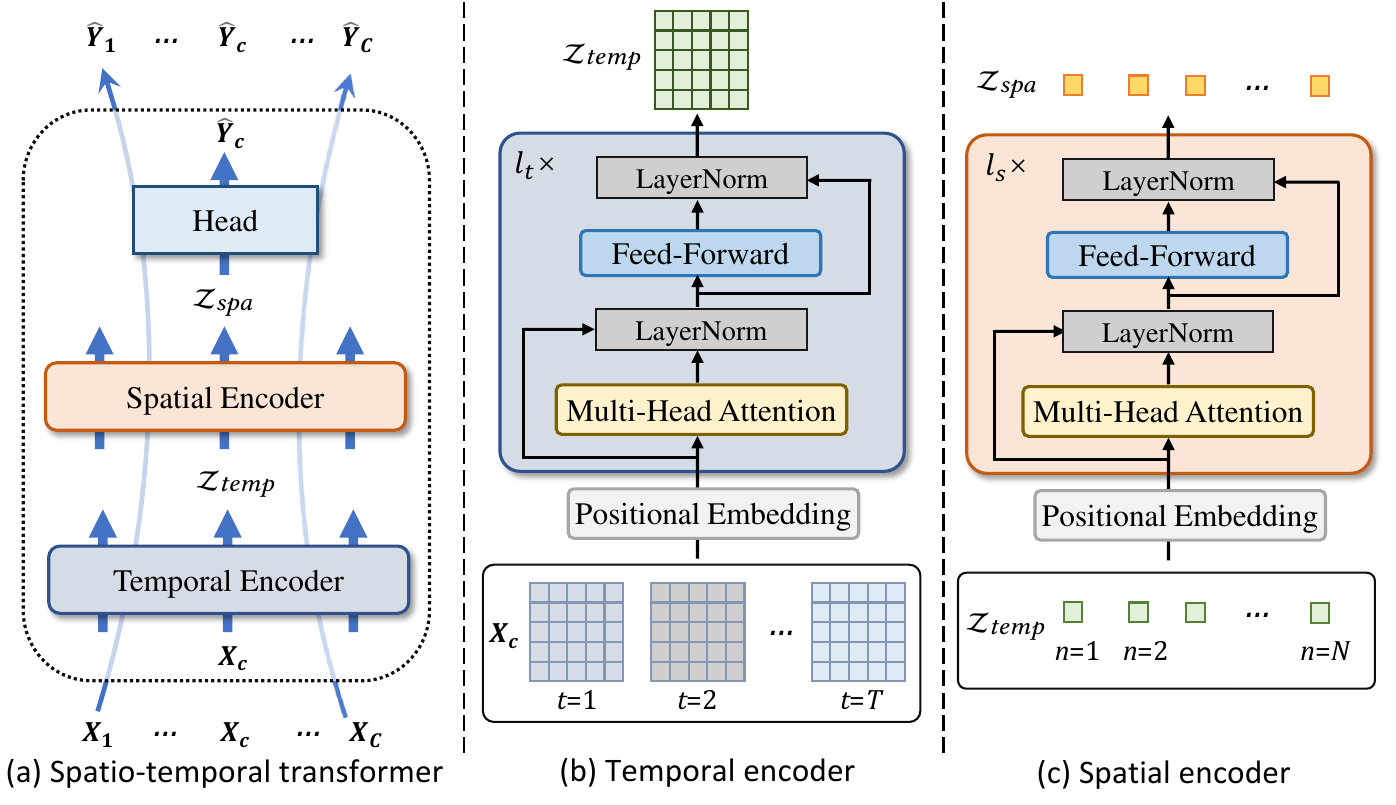}
% \vspace*{-4mm}
	\caption{
 {Pretrain phase of \name.}
 % Architecture of spatio-temporal transformer.
 (a) Our spatio-temporal transformer consists of temporal encoder, spatial encoder, and MLP-based head. It feeds on multiple spatio-temporal attributes in parallel.
 (b) The temporal encoder captures the temporally sequential relationship of $\boldsymbol{{X}}_c$ with positional embedding and $l_t$ layers of multi-head attention along with a feed-forward layer. It intercepts the feature map of the last timestep as output, \ie $\boldsymbol{\mathcal{Z}}_{temp}$.
 (c) The spatial encoder addresses the spatial information of $\boldsymbol{\mathcal{Z}}_{temp}$ with positional embedding as well as $l_s$ layers, and outputs spatial representation $\boldsymbol{\mathcal{Z}}_{spa}$.
 }
	\label{fig:model}
% 	\vspace*{-5mm}
\end{figure*}

\section{Preliminary}

% \emph{Definition 2.1.}
% \textbf{Region.}
% To describe timely social activities in a city particularly, we partition a city into $I\times J$ non-overlapping grids with equal size along with longitude and latitude \cite{strn, zjb_region}.
% Each grid represents a region in the city, let $\boldsymbol{x}_{t,c}\in \mathbb{R}^{N}$ represent the spatio-temporal attribute $c$ of the $N$ grids at timestep $t$, $N=I\times J$.
% coordinate in complex ways.

% \noindent\emph{Definition 2.2.}
% \textbf{Spatio-Temporal Multi-Attribute Prediction.}
% Given the spatio-temporal attribute $c$ of historical $T$ timesteps $\boldsymbol{x}_{1:T,c}=[\boldsymbol{x}_{1,c},\ldots,\boldsymbol{x}_{t,c},\ldots,\boldsymbol{x}_{T,c}]$ $\in \mathbb{R}^{T\times N}$, the goal of spatio-temporal prediction is to predict the future attribute.

% % Considering the common spatio-temporal pattern and dependency among different spatio-temporal attributes, 
% In this paper, we focus on spatio-temporal multi-attribute prediction, which aims to predict multiple spatio-temporal attributes of the future $H$ timesteps simultaneously.

\textbf{Spatio-Temporal Multi-Attribute Prediction.}
Let $\boldsymbol{x}_{1:T,c}=[\boldsymbol{x}_{1,c},$ $\ldots,\boldsymbol{x}_{t,c},\ldots,\boldsymbol{x}_{T,c}]$ $\in \mathbb{R}^{T\times N}$ represent the spatio-temporal attribute $c$ of historical $T$ timesteps and $N$ regions.
Given $C$ historical spatio-temporal attributes, spatio-temporal multi-attribute prediction aims to predict $C$ attributes of the future $H$ timesteps simultaneously.
Mathematically, the prediction could be formulated as follows,
\begin{equation}\label{Equ:Defination}
    \boldsymbol{x}_{1:T,c} \stackrel{f_{\boldsymbol{W}}}{\longrightarrow} \boldsymbol{y}_{T+1:T+H,c}, \forall c=1,\ldots,C
\end{equation}
where $f_{\boldsymbol{W}}$ parameterized by $\boldsymbol{W}$ represents the trainable model. $\boldsymbol{y}_{T+1:T+H,c} \in \mathbb{R}^{H\times N}$ denotes the spatio-temporal attribute $c$ of future $H$ timesteps. For a clear description, we denote $\boldsymbol{x}_{1:T,c}$ with $\boldsymbol{X}_c$, and let $\boldsymbol{\mathcal{X}}=[\boldsymbol{X}_1, \ldots, \boldsymbol{X}_c, \ldots, \boldsymbol{X}_C] \in \mathbb{R}^{T\times N\times C}$ represent the observed $C$ spatio-temporal attributes. 
Let $\boldsymbol{\mathcal{Y}} \in \mathbb{R}^{H\times N\times C}$ represent the future $C$ spatio-temporal attributes of $H$ timesteps.

% Considering multiple spatio-temporal attributes, the $D$ spatio-temporal attributes at timestep $t$ could be noted as $\boldsymbol{X}_t=[x_{t,1},\ldots,x_{t,d},\ldots,x_{t,D}]$ $\in \mathbb{R}^{N\times D}$.
% Given the historical spatio-temporal attributes of $T$ timesteps $\boldsymbol{{X}}_{1:T}$, spatio-temporal multivariate prediction aims to predict multiple spatio-temporal attributes of the future $H$ timesteps simultaneously.
% Mathematically, it could be formulated as follows,
% \begin{equation}\label{Equ:Defination}
%     \boldsymbol{X}_{1:T,d} \stackrel{f_{\boldsymbol{W}}}{\longrightarrow} \boldsymbol{y}_{T+1:T+H,d}, \forall d=1,\ldots,D
% \end{equation}
% where 

\section{Methodology}
{
Extensive efforts have been exerted in spatio-temporal prediction of a single attribute \cite{gwnet, strn, convgcn}.
However, these methods cannot well address spatio-temporal multi-attribute prediction:
Simply expanding the feature dimension to support multi-attribute ignores and cannot utilize the correlation among different attributes.
Training on multiple attributes separately requires tremendous computational and storage costs, which is impractical in application.
}

In this section, we present {a pretrain-prompt tuning scheme to address spatio-temporal multi-attribute prediction, named \name.
We first propose our benchmark architecture spatio-temporal transformer, and then introduce the pretrain and prompt tuning phases, respectively.
We detail our novel spatio-temporal prompt tokens, which fit the specific attributes with trivial trainable parameters.}

\subsection{Framework Overview}
We first propose an effective transformer-based architecture to capture common knowledge of multiple spatio-temporal attributes.
As shown in {Figure} \ref{fig:model} (a), from bottom to top, our transformer architecture consists of temporal encoder, spatial encoder, and head. 
% Basically, the three modules are in charge of temporal dependency capture, spatial dependency capture, and spatio-temporal state mapping.
To balance the model training on multiple attributes, we feed the transformer with single spatio-temporal attributes in parallel which share the model parameters. 
The temporal encoder and spatial encoder are illustrated in Figure \ref{fig:model} (b) and (c), respectively.

In the pretrain phase, we train the architecture in a parameter-sharing way so that the model learns the common spatio-temporal characteristics among all attributes.
To enhance the performance on specific spatio-temporal attributes, we propose a prompt tuning phase to fit each attribute, which benefits the specific attribute modeling from the learned common spatio-temporal patterns.
Specifically, by fixing the main body of the pretrained framework, \ie temporal encoder and spatial encoder, and only updating the prompt tokens and head, our \name can capture the specific spatial and temporal patterns of every single attribute with trivial trainable parameters.
In this way, our method can address common spatio-temporal characteristics shared among all spatio-temporal attributes, and fits specific ones in a parameter-efficient way.

\subsection{Spatio-Temporal Transformer} 
{
Due to the superior capacity in related tasks \cite{transformer, informer}, transformer has been introduced in spatio-temporal prediction \cite{traffictmr, tmr_icml22, tmr_tkde22, tmr_tnnls22}.
However, there is still a lack of a solution for spatio-temporal multi-attribute prediction, which can address the complex relationships among numerous attributes effectively.
In this section, we propose a swift transformer-based backbone which enjoys a compact architecture and compelling efficacy in modeling spatio-temporal dependency of multiple attributes. 
}
\label{subsec:sttransformer}
\subsubsection{Temporal Encoder}
Temporal dependency capture plays a prominent role in spatio-temporal prediction. 
In general, our temporal encoder consists of a stacked multi-head attention block and a position-wise feed-forward layer.

\noindent \textbf{Positional Embedding.}
In particular, we first employ an MLP-based mapping function to learn the embedding of each input spatio-temporal attribute $\boldsymbol{{X}}_c$ as $\boldsymbol{\mathcal{Z}}$. % \in \mathbb{R}^{T\times N\times D}$.
To learn the temporal information, we transpose the dimension of region and timestep of $\boldsymbol{{X}}_c$, \ie from $\mathbb{R}^{T\times N}$ to $\mathbb{R}^{N\times T}$, so 
the mapping operation is,
\begin{equation}\label{Equ:bottom}
    \boldsymbol{\mathcal{Z}} = \sigma(\boldsymbol{W}_m {\rm Trans}(\boldsymbol{{X}}_c)+\boldsymbol{b}_m)
\end{equation}
where $\boldsymbol{W}_m \in \mathbb{R}^{1\times D}$ and $\boldsymbol{b}_m \in \mathbb{R}^{D}$ are learnable weight and bias, respectively. $\boldsymbol{\mathcal{Z}} \in \mathbb{R}^{N\times T\times D}$ is the learned representation, and $D$ is the embedding size. $\sigma(\cdot)$ and ${\rm Trans(\cdot)}$ are sigmoid and transpose functions, respectively.

To integrate sequential information into the transformer, we utilize trainable positional embedding to mark the temporal sequence. Specifically, we add positional embedding $\boldsymbol{p}_{temp} \in \mathbb{R}^{T\times D}$ to $\boldsymbol{\mathcal{Z}}$, 

\begin{equation}\label{Equ:temporal_position}
    \boldsymbol{\mathcal{Z}}_{temp}^{(0)} = \boldsymbol{\mathcal{Z}}+\boldsymbol{p}_{temp}
\end{equation}

\noindent\textbf{Multi-Head Attention.}
The self-attention mechanism \cite{transformer} is the key component of transformer, it calculates the self-attention based on scaled dot-product, and outputs the attention-weighted sum of the feature.  
The self-attention calculation is shown as follows, 

\begin{equation}
     {\rm Att }(\boldsymbol{Q}, \boldsymbol{K}, \boldsymbol{V})={\rm softmax}\left(\frac{\boldsymbol{Q} \boldsymbol{K}^T}{\sqrt{D_k}}\right) \boldsymbol{V} 
\end{equation}
where $\boldsymbol{Q}$, $\boldsymbol{K}$, and $\boldsymbol{V}$ are the query, key, and value matrix, respectively.

The multi-head attention splits the input feature into several partitions, and conducts self-attention individually. Hence, the feature could be handled in different embedding subspaces, and then concatenated to integrate diverse information,
\begin{equation}
\begin{aligned}
\label{Equ:MHAtt}
{\rm MHAtt}(\boldsymbol{Q}, \boldsymbol{K}, \boldsymbol{V}) & ={\rm Concat}\left({head}_1, \ldots, {head}_{h}\right) \boldsymbol{W}^{\boldsymbol{O}} \\
 {head}_{{i}} & ={\rm Att}\left(
\boldsymbol{Q} \boldsymbol{W}_i^{\boldsymbol{Q}},
\boldsymbol{K} \boldsymbol{W}_i^{\boldsymbol{K}}, 
\boldsymbol{V} \boldsymbol{W}_i^{\boldsymbol{V}}
\right)
\end{aligned}
\end{equation}
where $\boldsymbol{W}_i^{\boldsymbol{Q}} \in \mathbb{R}^{D\times D_k}$, $\boldsymbol{W}_i^{\boldsymbol{K}} \in \mathbb{R}^{D\times D_k}$, $\boldsymbol{W}_i^{\boldsymbol{V}} \in \mathbb{R}^{D\times D_v}$ and $\boldsymbol{W}^O \in \mathbb{R}^{hD_v\times D}$ are learnable parameter matrices. 

Equipped with residual connection \cite{residual} and layer normalization \cite{layernorm}, the multi-head attention operation is conducted as,
\begin{equation}
\label{Equ:MX}
{{M}}(\boldsymbol{X}) ={\rm LayerNorm}(\boldsymbol{X}+{\rm MHAtt}(\boldsymbol{X},\boldsymbol{X}, \boldsymbol{X}))
\end{equation}

\noindent\textbf{Feed-Forward Layer.}
Then, position-wise feed-forward is applied to each position separately, which is followed by residual connection and layer normalization,
\begin{equation}
\begin{aligned}
\label{Equ:FX}
     {\rm FeedFwd}(\boldsymbol{X})&=\boldsymbol{W}_2{\rm ReLU}(\boldsymbol{W}_1\boldsymbol{X}+\boldsymbol{b}_1)+\boldsymbol{b}_2\\
     { F}(\boldsymbol{X})&={\rm LayerNorm}(\boldsymbol{X}+{\rm FeedFwd}(\boldsymbol{X}))     
\end{aligned}
\end{equation}
where $\boldsymbol{W}_1$, $\boldsymbol{W}_2$ are parameter matrices, and $\boldsymbol{b}_1$, $\boldsymbol{b}_2$ are biases. 

Based on Eq. \eqref{Equ:MX} and \eqref{Equ:FX}, the temporal intermediate representation after the $l$-th iteration $\boldsymbol{\mathcal{Z}}_{temp}^{(l)}$ can be calculated as,
\begin{equation}
\label{Equ:TempEncoder}
\begin{aligned}
\boldsymbol{\mathcal{Z}}_{temp}^{(l)} = { F}({ M}(\boldsymbol{\mathcal{Z}}_{temp}^{(l-1)})) 
\end{aligned}
\end{equation}

We conduct $l_t$ temporal encoder layers of multi-head attention and feed-forward layer to capture the temporal dependency. 
Then, we take the intermediate representation of the last timestep as the temporal representation $\boldsymbol{\mathcal{Z}}_{temp} \in \mathbb{R}^{N\times D}$ of $\boldsymbol{X}_{c}$,
% \zxy{of what?}, 

\begin{equation}
\label{Equ:TempRepresentation}
\boldsymbol{\mathcal{Z}}_{temp} = \boldsymbol{\mathcal{Z}}_{temp}^{(l_t)}\Big|_{t=T}
\end{equation}

% The temporal representation $\boldsymbol{\mathcal{Z}}_{temp}=\boldsymbol{\mathcal{Z}}_{temp}^{l_e}$. 

% $\boldsymbol{}$

\subsubsection{Spatial Encoder}
After capturing the temporal dependency with temporal encoder, we address the spatial relationship among the grids. 
{
Without extra computational cost such as k-hop adjacency matrix \cite{traffictmr, tmr_icml22}, we incorporate a simple but effective spatial encoder to learn spatial information. 
}
% \zxy{
% challenges
% }
Generally, we add spatial positional information, and process with stacked multi-head attention and feed-forward layer.

\noindent \textbf{Positional Embedding.}
Similar to Eq. \eqref{Equ:temporal_position}, given temporal representation $\boldsymbol{\mathcal{Z}}_{temp} \in \mathbb{R}^{N\times D}$, we add learnable positional embedding $\boldsymbol{p}_{spa} \in \mathbb{R}^{N\times D}$ to learn the positional information among the grids.
% To further enhance the positional information learning, we apply sine and cosine functions of different frequencies as spatial position token, which marks the relative positional information in a lightweight way \cite{transformer, traffictmr}. 
\begin{equation}\label{Equ:spatial_position}
\begin{aligned}
    \boldsymbol{\mathcal{Z}}_{spa}^{(0)} = \boldsymbol{\mathcal{Z}}_{temp}+\boldsymbol{p}_{spa}%+{\rm PE}(\boldsymbol{\mathcal{Z}}_{temp}) 
\end{aligned}
\end{equation}
% In particular, we calculate the constant positional token for region $r$ by ${\rm PE}(r, 2i)$ and ${\rm PE}(r, 2i+1)$, where $i$ is the dimension, 
% \lqd{$D$ is what? Besides, you have used the symbol in previous of the paper.}
% \begin{equation}
% \begin{aligned}
%     {\rm PE}(r, 2i)&=sin(r/10000^{2i/D})\\
%     {\rm PE}(r, 2i+1)&=cos(r/10000^{2i/D})
% \end{aligned}
% \end{equation}

\begin{figure}[!t]
\centering
	%	\hspace*{-7mm}1.049
	\includegraphics[width=0.7\linewidth]{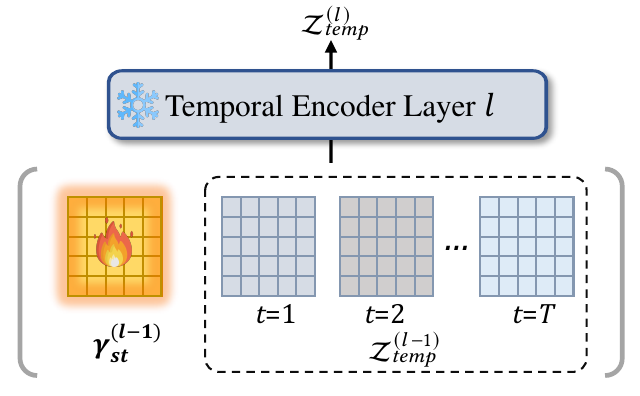}
	\caption{{Prompt tuning phase of \name. 
  \name keeps the backbone frozen and updates the prompt tokens and head. 
  The luminous grids $\boldsymbol{\gamma}_{st}$ represent the prompt tokens, and the gray grids $\boldsymbol{\mathcal{Z}}_{temp}$ denote the intermediate representation in the temporal encoder. 
  \name integrates prompt tokens to each layer to the input sequence of the temporal encoder. }
  % Tiny spatio-temporal prompt inserts prompt tokens to feature {(b)} sequence of temporal encoder, and (c) sequence of spatial encoder.
 }
    % \vspace{-3mm}
	\label{fig:prompt}
	% \vspace*{-3mm}
\end{figure}

\noindent \textbf{Spatial Representation Learning.}
The spatial encoder consists of stacked multi-head attention {${ M}(\cdot)$} and feed-forward layer {${ F}(\cdot)$}, as shown in Eq. \eqref{Equ:MX} and \eqref{Equ:FX}.
The spatial intermediate representation output by the $l$-th iteration $\boldsymbol{\mathcal{Z}}_{spa}^{(l)}$ is, 
\begin{equation}
\label{Equ:SpaEncoder}
\begin{aligned}
\boldsymbol{\mathcal{Z}}_{spa}^{(l)} = { F}({ M}(\boldsymbol{\mathcal{Z}}_{spa}^{(l-1)})) 
\end{aligned}
\end{equation}

Processed by $l_s$ layers of multi-head attention and feed-forward layer as spatial encoder, the spatial representation $\boldsymbol{\mathcal{Z}}_{spa} \in \mathbb{R}^{N\times D}$ is the output of the spatial encoder, 
% \zxy{what is the physical sense ?}
\begin{equation}
\label{Equ:SpaRepresentation}
\boldsymbol{\mathcal{Z}}_{spa} = \boldsymbol{\mathcal{Z}}_{spa}^{(l_s)}
\end{equation}

\subsubsection{Head}
To map the learned spatio-temporal representation to the future $H$ timesteps, we introduce an MLP-based head for the dimension transformation.
The predicted spatio-temporal attribute $c$'s future State $\boldsymbol{\hat{Y}}_c \in \mathbb{R}^{N\times H}$ can be calculated as,
\begin{equation}\label{Equ:head}
    \boldsymbol{\hat{Y}}_c = \sigma(\boldsymbol{W}_h\boldsymbol{\mathcal{Z}}_{spa}+\boldsymbol{b}_h)
\end{equation}
where $\boldsymbol{W}_h \in \mathbb{R}^{D\times H}$ and $\boldsymbol{b}_h \in \mathbb{R}^{H}$ are weight matrix and bias. 

% Appendix
% To highlight our spatio-temporal transformer, we present the pseudo-code in Algorithm \ref{alg:algorithm1} in \textbf{{Appendix \ref{subsec:alg}}}.

% \zxy{do we need a discussion of our difference with traditional ST transformers?}

\subsection{Pretrain Phase}

% \zxy{why pretrain?}
% \zxy{Inspired by the successful application of LLMs to maintain the common knowledge in diverse tasks \cite{ofa_pmt, gpt3, yan2022towards}, we devise a parameter-sharing training strategy to address the common characteristics among multiple spatio-temporal attributes.}
To address the common characteristics among multiple spatio-temporal attributes, we devise a parameter-sharing training strategy  \cite{ofa_pmt, gpt3, yan2022towards}.
% To address the common spatio-temporal knowledge, we train the benchmark with multiple spatio-temporal attributes in parallel.
In this pretrain phase, we train the spatio-temporal transformer with all the $C$ spatio-temporal attributes.
Specifically, we feed $C$ attributes in parallel, which share the same parameters as shown in Figure \ref{fig:model} (a).

Given the spatio-temporal attributes matrix $\boldsymbol{\mathcal{X}}=[\boldsymbol{X}_1, \ldots, \boldsymbol{X}_c,$ $ \ldots, \boldsymbol{X}_C] \in \mathbb{R}^{T\times N\times C}$, we feed spatio-temporal transformer with single spatio-temporal attribute matrix $\boldsymbol{X}_c \in \mathbb{R}^{T\times N}$ with $c \in [1,\ldots, C]$ simultaneously.
For the output $\boldsymbol{\hat{Y}}_c \in \mathbb{R}^{N\times H}$, we concatenate $C$ attributes as $\boldsymbol{\mathcal{\hat{Y}}}=[\boldsymbol{\hat{Y}}_1, \ldots, \boldsymbol{\hat{Y}}_c, \ldots, \boldsymbol{\hat{Y}}_C] \in \mathbb{R}^{H\times N\times C}$.

We denote the model parameters of temporal encoder and spatial encoder as $\boldsymbol{\theta}$, the parameters of head as $\boldsymbol{\omega}$, and the sum of the two constitute the spatio-temporal transformer.
% , \ie $\boldsymbol{W} = \{\boldsymbol{\theta}, \boldsymbol{\omega}\}$.

The optimization objective of pretrain could be formulated as,
\begin{equation}
\label{Equ:pretrain}
\begin{aligned}
% \min _{\boldsymbol{\theta},\boldsymbol{\omega}}
&\min _{\boldsymbol{\theta},\boldsymbol{\omega}} \mathcal{L}(\boldsymbol{\mathcal{X}}, \boldsymbol{\mathcal{{Y}}}; \boldsymbol{\theta},\boldsymbol{\omega}) :=\\
&\min _{\boldsymbol{\theta},\boldsymbol{\omega}} 
{\rm RMSE}( f_{\boldsymbol{\theta},\boldsymbol{\omega}}(\boldsymbol{\mathcal{X}}),
\boldsymbol{\mathcal{Y}})+
{\rm MAE}(f_{\boldsymbol{\theta},\boldsymbol{\omega}}(\boldsymbol{\mathcal{X}}),\boldsymbol{\mathcal{Y}})
% \boldsymbol{W}
% \\
% &\mathcal{L}(\boldsymbol{\mathcal{Y}}, \boldsymbol{\mathcal{\hat{Y}}}; \boldsymbol{\theta},\boldsymbol{\omega}) := RMSE(\boldsymbol{\mathcal{Y}}, \boldsymbol{\mathcal{\hat{Y}}})+MAE(\boldsymbol{\mathcal{Y}}, \boldsymbol{\mathcal{\hat{Y}}})
\end{aligned}
\end{equation}
where the model output $f_{\boldsymbol{\theta},\boldsymbol{\omega}}(\boldsymbol{\mathcal{X}}) =\boldsymbol{\mathcal{\hat{Y}}}$.
% Appendix
% The pseudo-code of pretrain phase is shown in Algorithm \ref{alg:algorithm2} in {\textbf{Appendix \ref{subsec:alg}}}.
% By jointly learning multiple attributes, the spatio-temporal transformer is able to learn the common spatio-temporal patterns among different attributes.

\subsection{Prompt Tuning Phase}

% \subsubsection{\zzj{Motivation}}

% \zxy{
% Try to move these two paragraphs of challenges into 3.4.1 and 3.4.2...
% }
Given the diverse attributes to be tackled in smart city construction, it demands a swift training strategy to save the training and deployment costs as much as possible.
In this section, we propose a pretrain-prompt tuning pattern to alleviate the heavy training cost while achieving better performance. 
We present \textit{\stpmt} and
% and its lightweight version \textit{\apmt}, 
by prompt tuning with it, we can well-fit the specific spatio-temporal pattern of each spatio-temporal attribute with trivial trainable parameters.

\subsubsection{\stpmthead}
During the pretrain phase, we obtain common spatio-temporal characteristics in $f_{\boldsymbol{\theta}, \boldsymbol{\omega}}$.
To enhance the approximation of each attribute, a straightforward solution is fine-tuning \cite{ft}, which loads and re-trains the whole model on single attributes.
However, pretrain fine-tuning pattern demands tremendous computational costs for tuning all backbone parameters, and suffers from catastrophic forgetting \cite{ft_forget, ft_forget_origin}.

To prompt the pretrained model with specific spatio-temporal attributes, we propose a novel \stpmt token into the input sequence. We build trainable \stpmt tokens with the same shape of intermediate temporal features, \ie $\boldsymbol{\gamma}_{st} \in \mathbb{R}^{N \times D}$. Then, we insert prompt tokens to the feature sequence of each layer of the temporal encoder so as to trigger the learned common spatio-temporal pattern in $f_{\boldsymbol{\theta}, \boldsymbol{\omega}}$.

Through trainable prompt tokens and optimizing with backbone tuning, this prompt tuning strategy is able to enhance the performance of the specific attributes based on the maintained common knowledge. 
The \stpmt tuning process is shown in Figure \ref{fig:prompt}.

Specifically, we integrate $n_{st}$ learnable \stpmt tokens $\boldsymbol{\gamma}_{st}^{(l-1)} \in \mathbb{R}^{N \times n_{st} \times D}$ to the input temporal representation of each layer $\boldsymbol{\mathcal{Z}}_{temp}^{(l-1)} \in \mathbb{R}^{N \times T \times D}$ on the temporal dimension.
Hence, the dimension of the concatenated temporal representation is
% $\boldsymbol{\mathcal{Z}}_{temp}^{(l-1)} \in$
$ \mathbb{R}^{N \times T+n_{st} \times D}$. 
After processing with multi-head attention and feed-forward layer, we truncate the added $\boldsymbol{\gamma}_{st}^{(l-1)}$ to keep the original sequence length, and the truncated $\boldsymbol{\mathcal{Z}}_{temp}^{(l)} \in \mathbb{R}^{N \times T \times D}$.
So the temporal intermediate representation in Eq. \eqref{Equ:TempEncoder} becomes as, 
% \begin{equation}
% \label{Equ:pmtTempEncoder}
% \begin{aligned}
% \boldsymbol{\mathcal{Z}}_{temp}^{(l-1)} &= Concat(\boldsymbol{\gamma}_{st}, \boldsymbol{\mathcal{Z}}_{temp}^{(l-1)}) \\
% \boldsymbol{\mathcal{Z}}_{temp}^{(l)} &= F(M(\boldsymbol{\mathcal{Z}}_{temp}^{(l-1)})) \\
% \boldsymbol{\mathcal{Z}}_{temp}^{(l)} &= Trunc(\boldsymbol{\mathcal{Z}}_{temp}^{(l)}, \boldsymbol{\gamma}_{st})
% \end{aligned}
% \end{equation}
\begin{equation}
\label{Equ:stpmtTempEncoder}
\begin{aligned}
\boldsymbol{\mathcal{Z}}_{temp}^{(l)} = {\rm Trunc}(F(M( {\rm Concat}(\boldsymbol{\gamma}_{st}^{(l-1)}, \boldsymbol{\mathcal{Z}}_{temp}^{(l-1)})))) \\
\end{aligned}
\end{equation}

\subsubsection{Prompt Tuning}
% [Challenge and benefits]
Recently, prompt learning has shown its effects in triggering the common knowledge maintained in the pretrained backbone \cite{lama, gpt3} and enhancing the target task performance. 
It integrates lightweight prompt tokens in the input sequence and tunes the pretrained backbone to prompt the backbone with the specific information of the target task. 
However, in spatio-temporal prediction, we need to prompt the backbone with the target task's temporal and spatial information, which is intricate to be defined compactly.
In this context, we elaborate trainable \stpmt tokens, to indicate both spatial and temporal properties of target attributes.

Given the pretrained $f_{\boldsymbol{\theta}, \boldsymbol{\omega}}$, rather than tuning the whole backbone parameters $f_{\boldsymbol{\theta}, \boldsymbol{\omega}}$ like fine-tuning,
the prompt tuning phase aims to tune a small part of the spatio-temporal transformer, \ie $f_{\boldsymbol{\gamma}_{st}, \boldsymbol{\omega}}, \boldsymbol{\gamma}= \boldsymbol{\gamma}_{st}$.
% \ie $f_{\boldsymbol{\gamma}, \boldsymbol{\omega}}, \boldsymbol{\gamma}\in \{\boldsymbol{\gamma}_{st}, \boldsymbol{\gamma}_{ti}\}$.
Specifically, we load $f_{\boldsymbol{\theta}, \boldsymbol{\omega}}$, freeze the $\boldsymbol{\theta}$, and initialize the $\boldsymbol{\omega}$ and prompt token $\boldsymbol{\gamma}$ {for each spatio-temporal attribute}. 
Mathematically, the optimizing process is,
\begin{equation}
\label{Equ:pretrain}
\begin{aligned}
&\min _{\boldsymbol{\gamma},\boldsymbol{\omega}} \mathcal{L}(\boldsymbol{\mathcal{X}}, \boldsymbol{\mathcal{{Y}}}; \boldsymbol{\gamma},\boldsymbol{\omega}) :=\\
&\min _{\boldsymbol{\gamma},\boldsymbol{\omega}} 
{\rm RMSE}( f_{\boldsymbol{\gamma},\boldsymbol{\omega}}(\boldsymbol{\mathcal{X}}),
\boldsymbol{\mathcal{Y}})+
{\rm MAE}(f_{\boldsymbol{\gamma},\boldsymbol{\omega}}(\boldsymbol{\mathcal{X}}),\boldsymbol{\mathcal{Y}})
\end{aligned}
\end{equation}
where the model output $f_{\boldsymbol{\gamma},\boldsymbol{\omega}}(\boldsymbol{\mathcal{X}}) =\boldsymbol{\mathcal{\hat{Y}}}$.
% We will verify the efficacy of \stpmt and \apmt, respectively.

% Appendix
% In Algorithm \ref{alg:algorithm2} in \textbf{Appendix \ref{subsec:alg}}, we summarize the pipeline of prompt tuning.

\subsubsection{{Discussion}}
{To comprehensively exhibit our prompt tuning strategy, we discuss several aspects in this section.}

\noindent \textbf{{Prompting Spatial and Temporal Information.}}
{
Prompt tuning aims to prompt the backbone with the target task's information by refined prompt tokens \cite{gpt3, lama}.
% Our \name integrates trainable prompt tokens and optimizes with model training.
}
It is noteworthy that by adding $\boldsymbol{\gamma}_{st}^{(l-1)} \in \mathbb{R}^{N \times n_{st} \times D}$ to each temporal encoder layer, our \name is able to capture the characteristic at both spatial and temporal level.
Specifically, the $n_{st}$ \stpmt tokens prompt the spatio-temporal transformer for the temporal information of downstream attribute. 
Besides, each \stpmt consists of $N$ tokens, which address the spatial information of each region.
{
So tuning \stpmt tokens is able to maintain both spatial and temporal characteristics of the target attribute and prompt the backbone.
}

\noindent \textbf{{Space Efficiency.}}
For $l_t$ layers in temporal encoder, we iteratively integrate and truncate the \stpmt tokens, so the total \stpmt tokens are $\boldsymbol{\gamma}_{st} \in \mathbb{R}^{l_t\times N \times n_{st} \times D}$. For example, given a common setting with $l_t=2, N=64, n_{st}=2,$ and $D=32$, $\boldsymbol{\gamma}_{st}$ owns 8,192 parameters.
Adding the head $\boldsymbol{\omega}$, prompt tuning of \stpmt trains 8,588 parameters, which is 11.17\% of the whole transformer.
{
That means, compared with fine tuning all backbone parameters, our \name reduces almost 89\% trainable parameters when tuning on target attributes.
Considering the numerous spatio-temporal attributes, one can pretrain the backbone once and prompt tune everywhere efficiently.
% [physical, transfer]
}

\noindent \textbf{{\name v.s. Existing Prompt Tuning Methods.}}
{
Compared with the existing prompt learning methods in NLP \cite{gpt3, lama, pmt_prefix, cdq_pmt} and CV \cite{vpt_pmt, ofa_pmt}, our technical contributions lie in two aspects.
We propose a novel backbone, \ie spatio-temporal transformer, and a corresponding pretrain strategy to address the common spatio-temporal pattern in multiple attributes.
Besides, we devise a novel \stpmt tuning strategy to indicate both spatial and temporal characteristics of target single attributes efficiently. 
% [Benefits of prompt tuning, difference between NLP methods]
}

% \zxy{do we need a discussion of our difference with traditional prompt design?}

% \begin{table}[]
% \begin{center}
% % \vspace{-1mm}
% 	\caption{Statistics of the datasets.
% }
% 	% \vspace{-3.1mm}
% 	\label{table:datainfo}
% 	\scalebox{0.95}{
% 	\begin{tabular}{@{}ccc@{}}
% 		% \toprule[1pt]
%   \toprule
% 		\textbf{Dataset} & \multicolumn{1}{c}{\textbf{Complaint}} & \textbf{NYC Taxi} \\ \midrule
% \# attributes & 19 & 4\\
% \midrule
% \# trajectories & 2.27 M & 14.09 M\\%165.11 M
% \midrule
% time span & 1/1/2013$\sim$31/12/2014 & 1/1/2014$\sim$31/3/2014 \\ % 1/1/2014$\sim$31/12/2014
% \midrule
% grid shape & (8, 8) & (10, 20)\\
% \midrule
% time interval & 3 h & 20 min \\ %1 h
% 		% \bottomrule[1pt]
% \bottomrule
% 	\end{tabular}}
% 	% \vspace{-2mm}
% \end{center}
% \end{table}

\begin{table*}[t]
\centering
\small
\caption{Overall experiment results. Best performances are bold, next best are underlined.
% ($\cdot$) represents the number of attributes with superior performance than fine-tune.
``\textbf{{\Large *}}'' indicates the statistically significant improvements (i.e., two-sided t-test with $p<0.05$) over the best baseline. 
``Parameter'' shows the number of trainable parameters on Complaint.
{``(a)'' means training methods on each attribute separately. ``(b)'' means training a single model for multiple attributes. ``(ours)'' represents the results of our method. }
}
\vspace{-1mm}
\scalebox{1.08}{
% \ADLnullwidehline
% \renewcommand{\arraystretch}{1.0}
\begin{tabular}{llccccr} 
\toprule
& \textbf{Dataset} & \multicolumn{2}{c}{Complaint} & \multicolumn{2}{c}{NYC Taxi} & \multirow{2}{*}{Parameter}\\ 
\cmidrule(lr){3-4}\cmidrule(lr){5-6}
& Metrics & RMSE & MAE & RMSE & MAE & \\

\cmidrule(lr){1-7}
\multirow{8}{*}{(a)}
% & ARIMA &  &  &  &  & - \\
& Conv-GCN & 1.6987 $\pm$ 0.0332 & 0.3205 $\pm$ 0.0076 & 38.9876 $\pm$ 0.0017 & 18.0373 $\pm$ 0.0025 & 848,960 \\
& HGCN & 1.5489 $\pm$ 0.0203 & 0.2945 $\pm$ 0.0047 & 22.6791 $\pm$ 0.3155 & 10.3204 $\pm$ 0.1961 & 593,307 \\
& ASTGCN & 1.4022 $\pm$ 0.0018 & \underline{0.2559 $\pm$ 0.0002} & 16.3039 $\pm$ 0.3518 & 7.4875 $\pm$ 0.1346 & 73,906 \\
& GWNet & 1.4426 $\pm$ 0.0075 & {0.2617 $\pm$ 0.0021} & 16.3151 $\pm$ 0.2104 & 7.2020 $\pm$ 0.0644 & 537,484 \\
& CCRNN & 1.7502 $\pm$ 0.0179 & 0.4609 $\pm$ 0.0398 & 62.3831 $\pm$ 4.9698 & 38.0159 $\pm$ 4.3689 & 292,855 \\ 
% & MTGNN & 1.1612 $\pm$ 0.0176 & 0.2243 $\pm$ 0.0018 & 13.7202 $\pm$ 0.1305 & 6.2788 $\pm$ 0.0487 & 200,572 \\
& GTS & 1.5902 $\pm$ 0.0037 & 0.2808 $\pm$ 0.0029 & 19.2680 $\pm$ 0.7675 & 8.4905 $\pm$ 0.0162 & 9,400,723 \\
& TrafficTmr & 1.7356 $\pm$ 0.0003 & 0.3210 $\pm$ 0.0010 & 60.6256 $\pm$ 3.8161 & 29.4076 $\pm$ 2.3559 & 220,108 \\
\cmidrule(lr){1-7}
\multirow{6}{*}{(b)}
& ASTGCN-Full & 1.7569 $\pm$ 0.0014 & 0.5536 $\pm$ 0.0002 & 54.8019 $\pm$ 0.2570 & 33.5280 $\pm$ 0.1283 & 91,150 \\
& GWNet-Full & 1.7489 $\pm$ 0.0087 & 0.3397 $\pm$ 0.0045 & 31.3766 $\pm$ 0.2506 & 13.9877 $\pm$ 0.2057 & 538,636 \\
& CCRNN-Full & 2.0607 $\pm$ 0.5015 & 0.8617 $\pm$ 0.5339 & 64.2304 $\pm$ 10.7221 & 41.8823 $\pm$ 9.0472 & 315,373 \\
% & MTGNN-Full & 1.7824 $\pm$ 0.0109 & 0.3643 $\pm$ 0.0126 & 36.2450 $\pm$ 0.6608 & 16.2401 $\pm$ 0.3480 & 576,700 \\
& TrafficTmr-Full & 1.8333 $\pm$ 0.0036 & 0.6158 $\pm$ 0.0074 & 67.5295 $\pm$ 1.7047 & 41.4542 $\pm$ 2.3701 & 227,812 \\
& DMSTGCN & 1.7585 $\pm$ 0.0002 & 0.3274 $\pm$ 0.0014 & 30.8563 $\pm$ 0.6408 & 14.0257 $\pm$ 0.1979 & 630,468 \\
% & PLE &  &  &  &  &\\
\cmidrule(lr){1-7}
\multirow{4}{*}{(ours)}
& Single-Train & 1.4730 $\pm$ 0.0004 & 0.2863 $\pm$ 0.0006 & \underline{14.4828 $\pm$ 0.0689} & 6.8693 $\pm$ 0.8456 & 76,876 \\%76,876 *19 = 1,460,644
& Full-Train & \underline{1.1673 $\pm$ 0.0124} & 0.3109 $\pm$ 0.0070 & 15.2143 $\pm$ 0.1999 & 7.0856 $\pm$ 0.1173 & 76,876 \\
& Fine-Tune & 1.2362 $\pm$ 0.0101 & {0.2653 $\pm$ 0.0016} & 14.5352 $\pm$ 0.0238 & \underline{6.7474 $\pm$ 0.0285} & 76,876 \\
% & \textbf{\name{\_ti}} & {0.2535} & {1.1759} & \zzj{7.1594} & \zzj{15.1837} & 652 \\%\textbf{\name$_{\boldsymbol{ti}}$} \textsubscript
%  (5) (13) 
& \textbf{\name} & \textbf{1.1189* $\pm$ 0.0027} & \textbf{0.2432* $\pm$ 0.0054} & \textbf{14.2907* $\pm$ 0.0738} & \textbf{6.6882* $\pm$ 0.0251}  & 8,588 \\
% (7) (14) (2) (2)
\bottomrule

\end{tabular}
\label{table:overall}}
\end{table*}

\section{Experiment}

In this section, we comprehensively verify the efficacy of \name against advancing baselines. 
We conduct further transferability experiments to investigate the potential on unseen spatio-temporal attributes.
Besides, we devise ablation study and hyper-parameter analysis to verify the contribution of key components. 
% We also present the visualization of prompt tokens.
We also present the time and space efficiency comparison, which further emphasizes the efficacy and effectiveness of our \name.

\subsection{Datasets}
We eavluate \name on two physical spatio-temporal multi-attribute datasets: \textbf{Complaint}\footnote{\url{https://opendata.cityofnewyork.us/}} and \textbf{NYC Taxi}\footnote{\url{https://www1.nyc.gov/site/tlc/about/tlc-trip-record-data.page}}. 
The datasets are split into training, validation, and test sets by the ratio of 7:1:2. 
% The dataset properties are listed in Table \ref{table:datainfo}.
% Appendix
% in \textbf{Appendix \ref{subsec:baselines}}. 

\noindent\textbf{Complaint.}
The NYC 311 service offers assistance to requests, and record the 216 types of complaint with corresponding location and time.
We collect records of 19 kinds of complaints with the largest volumes in 2013 and 2014 as dataset Complaint, including 2.27M complaints about noise, dirty conditions, traffic signal condition, illegal parking, \etc 
% Appendix
% The full type list is shown in \textbf{Appendix \ref{subsec:implementation}}. 

\noindent\textbf{NYC Taxi.}
NYC Taxi datasets record taxi trajectories in New York City. Each trajectory consists of trip information, including fare, number of passengers, start and arrival time and corresponding geological properties. 
To describe the timely flows of vehicles and passengers, We compute the traffic inflow, traffic outflow, population inflow, and population outflow between city regions as spatio-temporal attributes.

\subsection{Baselines}
We compare our \name with the advancing baselines from three lines:
(a) {Spatio-temporal prediction:} Train models on each single attribute individually. 
% \textbf{ARIMA} \cite{arima},
\textbf{Conv-GCN} \cite{convgcn},
\textbf{HGCN} \cite{hgcn},
\textbf{ASTGCN} \cite{astgcn},
\textbf{GWNet} \cite{gwnet},
\textbf{CCRNN} \cite{ccrnn},
% \textbf{MTGNN} \cite{mtgnn}, 
\textbf{GTS} \cite{gts},
and \textbf{TrafficTmr} \cite{traffictmr}.
(b) {Spatio-temporal multi-attribute prediction:} Expand feature dimension to multiple attributes and train a single model on multiple attributes. 
\textbf{ASTGCN-Full},% \cite{astgcn},
\textbf{GWNet-Full},% \cite{gwnet},
\textbf{CCRNN-Full},% \cite{ccrnn},
% \textbf{MTGNN-Full} \cite{mtgnn},
\textbf{TrafficTmr-Full},% \cite{traffictmr}
and \textbf{DMSTGCN} \cite{dmstgcn}.
% and \textbf{PLE} \cite{tang2020progressive}.
(ours) Our spatio-temporal transformer in three training strategies:
\textbf{Single-Train}: train spatio-temporal transformer on each attribute individually,
\textbf{Full-Train}: train spatio-temporal transformer on all attributes, 
and \textbf{Fine-Tune}: tune full-trained model on each attribute individually.
% Appendix
% Please find the detailed description and experimental setting of baselines in {\textbf{Appendix \ref{subsec:baselines}}}.

\subsection{Experiment Setups}
% We publicize the implementation code to ease reproducibility\footnote{\url{https://ufile.io/mu5wq4fn}}.
We predict the spatio-temporal attributes of future 12 timesteps based on historical 12 timesteps, \ie $T=H=12$.
To evaluate the model capacity, we use root mean squared error (RMSE) and mean absolute error (MAE) as evaluation metrics.
We incorporate two layers for the temporal encoder and spatial encoder for the spatio-temporal transformer, respectively, \ie $l_t = l_s = 2$. We integrate $n_{st} = 2$ prompt tokens in the tuning phase. We use Adam as optimizer with a learning rate of 0.003 and a hidden size of 32.
All experiments are conducted on one NVIDIA 2080-Ti GPU, and we compute the average result of 5 repetitions for all the results.
% Appendix
% The architecture and optimization details of \name could be referred to {\textbf{Appendix \ref{subsec:implementation}}}.
% \zxy{not only architecture details, but introduce the optimization(pre-train, prompt tune) details}.

\subsection{Overall Performance}
According to the experimental results in Table \ref{table:overall}, we can safely draw conclusions below,

% \begin{itemize}[leftmargin=*]
    % \item 
\textit{(1)} The advancing spatio-temporal prediction methods, \ie GraphWaveNet, CCRNN and MTGNN, achieve poor results on spatio-temporal multi-attribute prediction. Upgrading the feature dimension to fit multiple attributes can not handle the entangled spatio-temporal patterns among different attributes. Hence, the multi-attribute version of the two performs worse than the single-training ones, where the latter runs on each attribute repetitively.
\textit{(2)} The full-trained spatio-temporal transformer achieves a remarkably leading position against all the baselines, with a relatively small parameter scale. This verifies the compact and effective architecture of our spatio-temporal transformer and the efficacy of the parameter-sharing training strategy towards spatio-temporal multi-attribute prediction.
\textit{(3)} Single-Train achieves comparable results with Full-Train, which is in accordance with the expectation because in Single-Train the spatio-temporal transformer can well-fit the specific pattern of each attribute. However, the Full-Train defeats Single-Train in some scenarios, \ie RMSE of 1.1673 v.s. 1.4730 on Complaint. This indicates that there exists common spatio-temporal knowledge among different attributes, and well-exploiting the common pattern contributes to spatio-temporal prediction. 
% Our full-training spatio-temporal transformer is able to capture the common spatio-temporal knowledge among multiple attributes, which benefits the learning of each attribute.
\textit{(4)} According to the performance of Fine-Tune and Full-Train, fine-tuned transformer generally exceeds the full-trained one, except for the RMSE on Complaint. It could be ascribed to the catastrophic forgetting \cite{ft_forget, ft_forget_origin}, which erases the common spatio-temporal knowledge learned in the pretrained model and degenerates to fitting the spatio-temporal pattern of the single attribute as Single-Train.
\textit{(5)} We can observe a consistent improvement of \name beyond the fine-tuned result on all metrics of both datasets with about 10\% parameters tuned.  Specifically, \name achieves better RMSE results on 14 out of 19 spatio-temporal attributes in Complaint, RMSE results on 3 out of 4 spatio-temporal attributes in NYC Taxi. Our prompt tuning strategy alleviates the catastrophic forgetting phenomenon and well-fits the specific characteristic of the target spatio-temporal attribute.

Hence, our \name achieves promising performance on spatio-temporal multi-attribute prediction. 
Basically, our spatio-temporal transformer trained with the parameter-sharing strategy attains superior results compared with advanced baselines.
Besides, our prompt tuning strategy further addresses the spatio-temporal pattern of specific attributes, while maintaining the common knowledge of multiple attributes learned in pretrain stage.

\subsection{Transferability}
Considering the wide variety of tasks demanding to be dealt with in real-world city management scenarios, it is impractical to expand the data with new (unseen) attributes and retrain the model over and over again.
In this context, it is necessary to transfer the learned spatio-temporal knowledge to new attributes.
In this subsection, we investigate the key factor contributing to physical application, \textit{how does \name perform when tuning on new attributes?}

To answer this research question, we divide the Complaint dataset into two parts, \ie \textbf{ComplaintA} and \textbf{ComplaintB}, with 10 and 9 spatio-temporal attributes, respectively.
We devise experiments to simulate the scenario of tuning the pretrained spatio-temporal transformer on new spatio-temporal attributes. 
For example, 
we pretrain spatio-temporal transformer on ComplaintB, \ie source dataset, then prompt tune on ComplaintA, \ie target dataset, and report the result as ComplaintA in Table \ref{table:transfer}.
% In particular,
% we pretrain spatio-temporal transformer on the source dataset, and compare the results of fine-tuning and prompt tuning on the target dataset.
To position the performance, we present the results of single-train, full-train of spatio-temporal transformer on target dataset. We also pretrain and prompt-tune \name on target dataset, and denote as {\namewospace$\dagger$}.

From the results in Table \ref{table:transfer}, several conclusions can be made:
\textit{(1)} In accord with the expectation, \namewospace$\dagger$ achieves the best performance. Being pretrained and tuned on the target dataset, it achieves a good balance between the common spatio-temporal knowledge of all attributes and the concrete pattern of every single attribute in the target dataset.
\textit{(2)} Fine-tune and prompt tune get better results than full-train on some metrics, \eg MAE of both datasets of fine-tune, which shows there exists beneficial common information from source towards target attributes.
% Meanwhile, we also observe that 
\textit{(3)} Our \name with \stpmt attains the best performance in the transferring setting.
It freezes the main body of pretrained spatio-temporal transformer and addresses characteristics of target attributes with novel prompt tokens, which maintains the useful common knowledge and well-fits the target attributes unseen during pretraining.
It closely approaches the best performance and serves as a promising solution for physical scenario applications.

% \zxy{a brief summary, good transferability indicates...?}

% Similar to the overall results, full-train gets better results than single-train,

% \subsection{\zxy{No efficiency comparison? time or space}}

% \subsection{Ablation Study}

% Table \ref{table:ablation} shows the results of \name and the three variants.
% Based on the comparisons on Complaint, we make the following conclusions:
% \textit{(1)} The variant w/o prompt gets the worst performance, which demonstrates that adding prompt tokens benefits fitting the spatio-temporal pattern of target single attribute.
% On the other hand, we can observe that the performance is better than fine-tune. 
% The possible reason is that tuning partial parameters, \ie head $\boldsymbol{\omega}$, of pretrained transformer alleviates the catastrophic forgetting and transfers common knowledge towards target attribute \cite{partial_ft_aaai}.
% \textit{(2)} Shallow prompt only inserts \stpmt tokens to the first layer of temporal encoder in transformer and attains competitive results, which shows the capacity of \name about transferring common spatio-temporal knowledge to the target attribute. 
% \textit{(3)} As a natural way of integrating trainable tokens to input sequences of transformer, add prompt also achieves promising results. 
% The performance gap below our \name could be the latter's intuitive functionality of prompting the pretrained benchmark of the target task as part of the input sequence \cite{pmt_prefix, lpf_pmt}.

\subsection{{Tiny PromptST and Ablation Study}}
To further explore the efficacy of our \stpmt tokens and pursue the trade-off between performance and tuning parameter volume, we put forward a tiny version of \stpmt, \ie \apmt, which owns much less updated parameter volumes of \stpmt.
% \zxy{figure? otherwise hard to follow} 
Particularly, we insert \apmt tokens to both temporal encoder and spatial encoder in spatio-temporal transformer \ie 
$\boldsymbol{{\gamma}}_{ti} =[$
$\boldsymbol{\overline{\gamma}}_{ti} \in \mathbb{R}^{l_t\times n_{ti} \times D}$, $\boldsymbol{\ddot{\gamma}}_{ti} \in \mathbb{R}^{l_s \times n_{ti}\times D}]$.
% Appendix
% We illustrate the prompt tuning with \apmt in \textbf{Appendix \ref{subsec:tiny}}.

For temporal encoder, similar with \stpmt, we insert $n_{ti}$ learnable \apmt tokens $\boldsymbol{\overline{\gamma}}_{ti}^{(l-1)} \in \mathbb{R}^{n_{ti} \times D}$ to the input temporal representation of each layer $\boldsymbol{\mathcal{Z}}_{temp}^{(l-1)} \in \mathbb{R}^{N \times T \times D}$ on the temporal dimension, and the concatenated temporal representation is with shape of $\mathbb{R}^{N \times (T+n_{ti}) \times D}$.
It is worth noting that we expand the $\boldsymbol{\overline{\gamma}}_{ti}^{(l-1)}$ on $N$ grids, and then concatenate with temporal representation, so $\boldsymbol{\overline{\gamma}}_{ti}^{(l-1)}$ in temporal encoder concentrates on temporal sequential capture, and different grids in temporal representation are concatenated with the same token. 
The Eq. \eqref{Equ:TempEncoder} could be reformulated as,
\begin{equation}
\label{Equ:apmtTempEncoder}
\begin{aligned}
\boldsymbol{\mathcal{Z}}_{temp}^{(l)} = {\rm Trunc}(F(M( {\rm Concat}({\rm Expand}(\boldsymbol{\overline{\gamma}}_{ti}^{(l-1)}), \boldsymbol{\mathcal{Z}}_{temp}^{(l-1)})))) \\
\end{aligned}
\end{equation}

In addition, after adding spatial positional embedding, we insert $n_{ti}$ \apmt tokens $\boldsymbol{\ddot{\gamma}}_{ti}^{(l-1)} \in \mathbb{R}^{n_{ti} \times D}$ on the spatial representation $\boldsymbol{\mathcal{Z}}_{spa}^{(l-1)} \in \mathbb{R}^{N\times D}$ fed to each layer of the spatial encoder. 
The concatenated spatial representation owns shape of $\mathbb{R}^{(N+n_{ti})\times D}$. After each multi-head attention and feed-forward layer, we also truncate the added $\boldsymbol{\ddot{\gamma}}_{ti}^{(l-1)}$ to remain the same spatial sequence length. The Eq. \eqref{Equ:SpaEncoder} is changed as,
\begin{equation}
\label{Equ:apmtSpaEncoder}
\begin{aligned}
\boldsymbol{\mathcal{Z}}_{spa}^{(l)} = {\rm Trunc}(F(M( {\rm Concat}(\boldsymbol{\ddot{\gamma}}_{ti}^{(l-1)}, \boldsymbol{\mathcal{Z}}_{spa}^{(l-1)})))) \\
\end{aligned}
\end{equation}

For $l_t$ layers of temporal encoder and $l_s$ layers of spatial encoder, \name adds $\boldsymbol{\gamma}_{ti} \in \mathbb{R}^{(l_t+l_s) \times n_{ti} \times D}$ in total.
Following the former setting with $l_t=2, l_s=2, n_{ti}=2,$ and $D=32$, $\boldsymbol{\gamma}_{ti}$ only contains 256 parameters, \ie 0.33\% of the benchmark transformer.
With the parameters of head $\boldsymbol{\omega}$, \apmt tuning only requires 652 trainable parameters, which is less than \textbf{1\%} of the transformer $f_{\boldsymbol{\theta},\boldsymbol{\omega}}$. 

To comprehensively verify its efficacy, 
we also present several variants of our \name in this subsection to illustrate the contribution of each component.

\begin{table}[!t]
\centering
\caption{Transferability performance of \name. Best performances are bold and
\vspace{-1mm}
($\cdot$) represents the number of attributes with superior performance than fine-tune. 
\namewospace$\dagger$ presents \name pretrained and prompt tuned with \stpmt on the target dataset.
}
\resizebox{1\linewidth}{!}{%
% \ADLnullwidehline
% \renewcommand{\arraystretch}{1.0}
\begin{tabular}{lcccc} 
\toprule
\multicolumn{1}{l}{\textbf{Dataset}} & \multicolumn{2}{c}{{ComplaintA}} & \multicolumn{2}{c}{{ComplaintB}} \\ 
\cmidrule(lr){2-3}
\cmidrule(lr){4-5}
Metrics & MAE & {RMSE} & {MAE} & {RMSE} \\
\cmidrule(lr){1-5}

% \hline
% \multirow{2}{*}{(a)}
% & CCRNN &  &  &  &  &  \\
% & CCRNN-Full & 0.7487 & 2.0772 & 82.14 & 401.76 & 123,058 \\
% & CCRNN & 0.5567 & 1.9376 & 51.06 & 285.42 & 123,058 \\
% & MTGNN-Full & 0.5191 & 2.0259 & 25.79 & 210.63 & 362,620 \\
% & MTGNN & 0.3269 & 1.3046 & 13.58 & 118.17 & 362,620 \\
% \hline
\textbf{\namewospace$\dagger$} & \textbf{0.1740} & \textbf{0.7099} & \textbf{0.3172} & \textbf{1.5670} \\
% \hline
\cmidrule(lr){1-5}
Single-Train & 0.1901 & 0.8216 & 0.3882 & 2.1872 \\
Full-Train & 0.1999 & 0.7307 & 0.4210 & {1.7591} \\
Fine-Tune & 0.1855 & {0.7141} & 0.3769 & 2.0120 \\
% \textbf{\name{\_ti}} & \underline{0.1847(4)} & 0.7173(1) & \underline{0.3400(5)} & \zzj{1.7891(4)} \\
\textbf{\name} & \textbf{0.1785(6)} & \textbf{0.7120(3)} & \textbf{0.3243(5)} & \textbf{1.6275(5)} \\
% \textbf{\name{\_st}} & \textbf{0.1785(6)} & \textbf{0.7120(3)} & \textbf{0.3243(5)} & \textbf{1.6275(5)} \\
% \hline
\bottomrule

\end{tabular}
\label{table:transfer}
}
    \vspace{-2mm}
\end{table}

\begin{table}[t]
\centering
\small
\caption{Components analysis of \name.}
    \vspace{-1mm}
\scalebox{1.2}{
% \ADLnullwidehline
% \renewcommand{\arraystretch}{1.0}
\begin{tabular}{lccr} 
% \hline
\toprule
\multicolumn{1}{l}{\textbf{Dataset}} & \multicolumn{2}{c}{{Complaint}} & \multirow{2}{*}
{{Parameter}}\\ 
\cmidrule(lr){2-3}
Metrics & MAE & {RMSE} &  \\

% \hline
\cmidrule(lr){1-4}
w/o prompt & 0.2584 & 1.1918 & 396 \\
shallow prompt & 0.2435 & 1.1271 & 4,492 \\
add prompt & 0.2434 & 1.1386 & 8,588 \\
\textbf{\name{\_ti}} & 0.2535 & 1.1759 & 652 \\
\textbf{\name} & \textbf{0.2432} & \textbf{1.1189} & 8,588 \\
% \hline
\bottomrule

\end{tabular}
\label{table:ablation}}
    \vspace{-2mm}
\end{table}

\begin{figure}[!t]
{\subfigure{\includegraphics[width=0.9\linewidth]{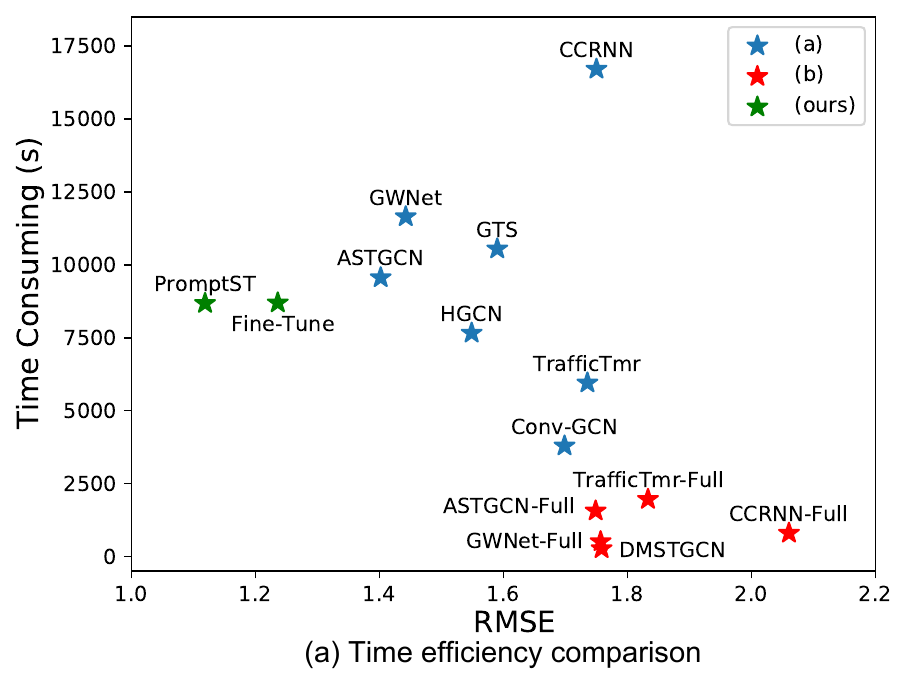}}}
\vspace{-3mm}

{\subfigure{\includegraphics[width=0.9\linewidth]{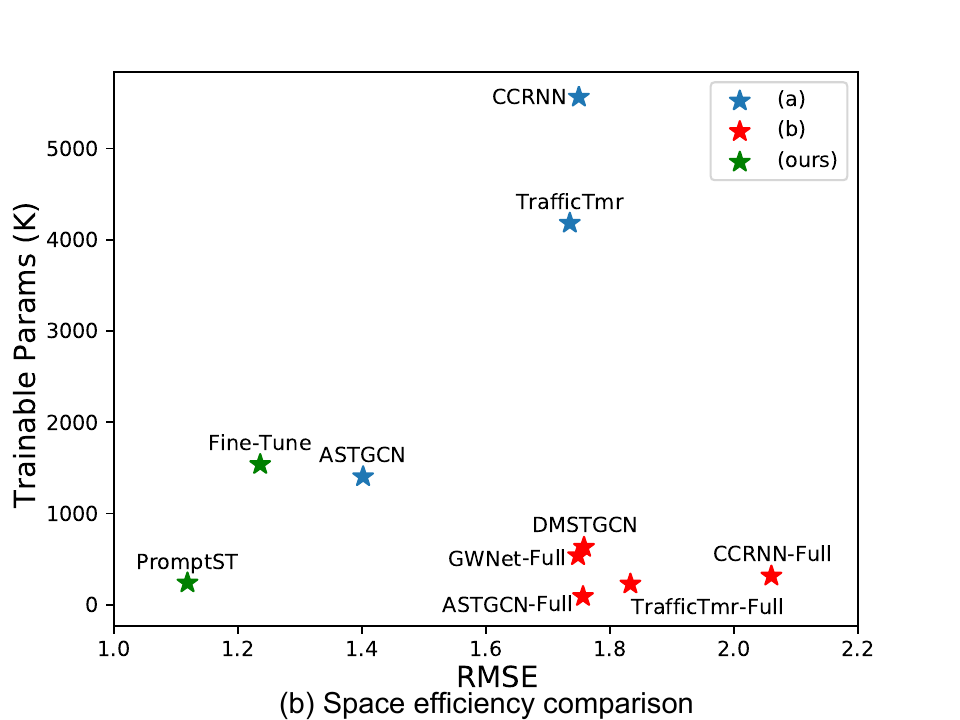}}}
    \vspace{-3mm}
    \caption{Time and space efficiency comparison. }
    \label{fig:time}
    \vspace{-3mm}
\end{figure}

\begin{itemize}[leftmargin=*]
    \item \textbf{tiny spatio-temporal prompt}: In the prompt tuning phase, we add $\boldsymbol{\gamma}_{ti}$ to the spatio-temporal transformer instead of $\boldsymbol{\gamma}_{st}$, and tune the prompt tokens and head.
\end{itemize}

\begin{itemize}[leftmargin=*]
    \item \textbf{w/o prompt}: After pretraining the spatio-temporal transformer $f_{\boldsymbol{\theta},\boldsymbol{\omega}}(\boldsymbol{\mathcal{X}})$, we fix the mainbody of transformer $\boldsymbol{\theta}$ and only tune the head $\boldsymbol{\omega}$, \ie $f_{\boldsymbol{\omega}}(\boldsymbol{\mathcal{X}})$, and verify the efficacy of prompt tokens.
\end{itemize}

\begin{itemize}[leftmargin=*]
    \item \textbf{shallow prompt}: Rather than integrating tokens to all the $l_t=2$ layers of temporal encoder, we only insert \stpmt token to the first layer.
\end{itemize}

\begin{itemize}[leftmargin=*]
    \item \textbf{add prompt}: Except for concatenating prompt tokens to input sequence of spatio-temporal transformer, we test another natural way, \ie adding the prompt tokens with each token in the input sequence repetitively.
\end{itemize}

Table \ref{table:ablation} shows the results of \name and the three variants.
Based on the comparisons on the Complaint, we make the following conclusions:
\textit{(1)} The variant w/o prompt gets the worst performance, which demonstrates that adding prompt tokens benefits fitting the spatio-temporal pattern of target single attribute.
On the other hand, we can observe that the performance is better than fine-tune. 
The possible reason is that tuning partial parameters, \ie head $\boldsymbol{\omega}$, of pretrained transformer alleviates the catastrophic forgetting and transfers common knowledge towards target attribute \cite{partial_ft_aaai}.
\textit{(2)} Shallow prompt only inserts \stpmt tokens to the first layer of temporal encoder in transformer and attains competitive results, which shows the capacity of \name about transferring common spatio-temporal knowledge to the target attribute. 
\textit{(3)} As a natural way of integrating trainable tokens to input sequences of transformer, adding prompt also achieves promising results. 
\textit{(4)} Our tiny version of prompt tuning, \apmt, also achieves competitive performance against the baselines with less than 1\% of benchmark and around 7\% parameters of \name. It addresses concrete temporal and spatial relationships by integrating lightweight prompt tokens into temporal encoder and spatial encoder separately, which is effective in fitting the target attributes. 

The performance gap below our \name could be the latter's intuitive functionality of prompting the pretrained benchmark of the target task as part of the input sequence \cite{pmt_prefix, lpf_pmt}.

\subsection{{Time and Space Efficiency}}
To evaluate our \name in the physical view, we compare the training time and total trainable parameters with the baselines, as shown in Figure \ref{fig:time}.
We present the performance on Complaint, which means for baselines in (a): training time and parameters are accumulations of individual training on 19 attributes; (b): results are from a single train on multi-attribute data; (ours) results consist of pretrain and individual tuning on 19 attributes.
We omit Conv-GCN, HGCN, GWNet, and GTS in Figure \ref{fig:time} (b) due to their tremendously large parameters volume, the concrete number of which can be referred to Table \ref{table:overall}.
According to the efficiency comparison, 
\textit{(1)} The advancing spatio-temporal prediction methods (blue stars) cost apparently more than their multi-attribute versions (red stars), while achieving consistently better results. 
\textit{(2)} As promising spatio-temporal multi-attribute solutions, fine-tuned spatio-temporal transformer and our \name take the relatively leading position.
\textit{(3)} Our \name achieves the best performance with trivial trainable parameters and comparable training hours.

% \begin{table}[h]
% \centering
% \caption{Components analysis of \name.}
% \scalebox{1.08}{
% % \ADLnullwidehline
% % \renewcommand{\arraystretch}{1.0}
% \begin{tabular}{c||cc||cc} 
% \hline
% \multicolumn{1}{c||}{\textbf{Dataset}} & \multicolumn{2}{c||}{{Complaint}} & \multicolumn{2}{c}{{NYC Taxi}}\\ 
% Metrics & MAE & {RMSE} & MAE & {RMSE} \\

% \hline
% w/o prompt & 0.2584 & 1.1918 & 7.0001 & 14.9954 \\
% shallow prompt & 0.2435 & 1.1271 & 6.8217 & 14.5773 \\
% add prompt & 0.2434 & 1.1386 & \textbf{6.7252} & 14.4478 \\
% \textbf{\name{\_st}} & \textbf{0.2432} & \textbf{1.1189} & 6.7384 & \textbf{14.4383} \\
% \hline

% \end{tabular}
% \label{table:ablation}}
% \end{table}

\begin{figure}[t]
    	\includegraphics[width=1\linewidth]{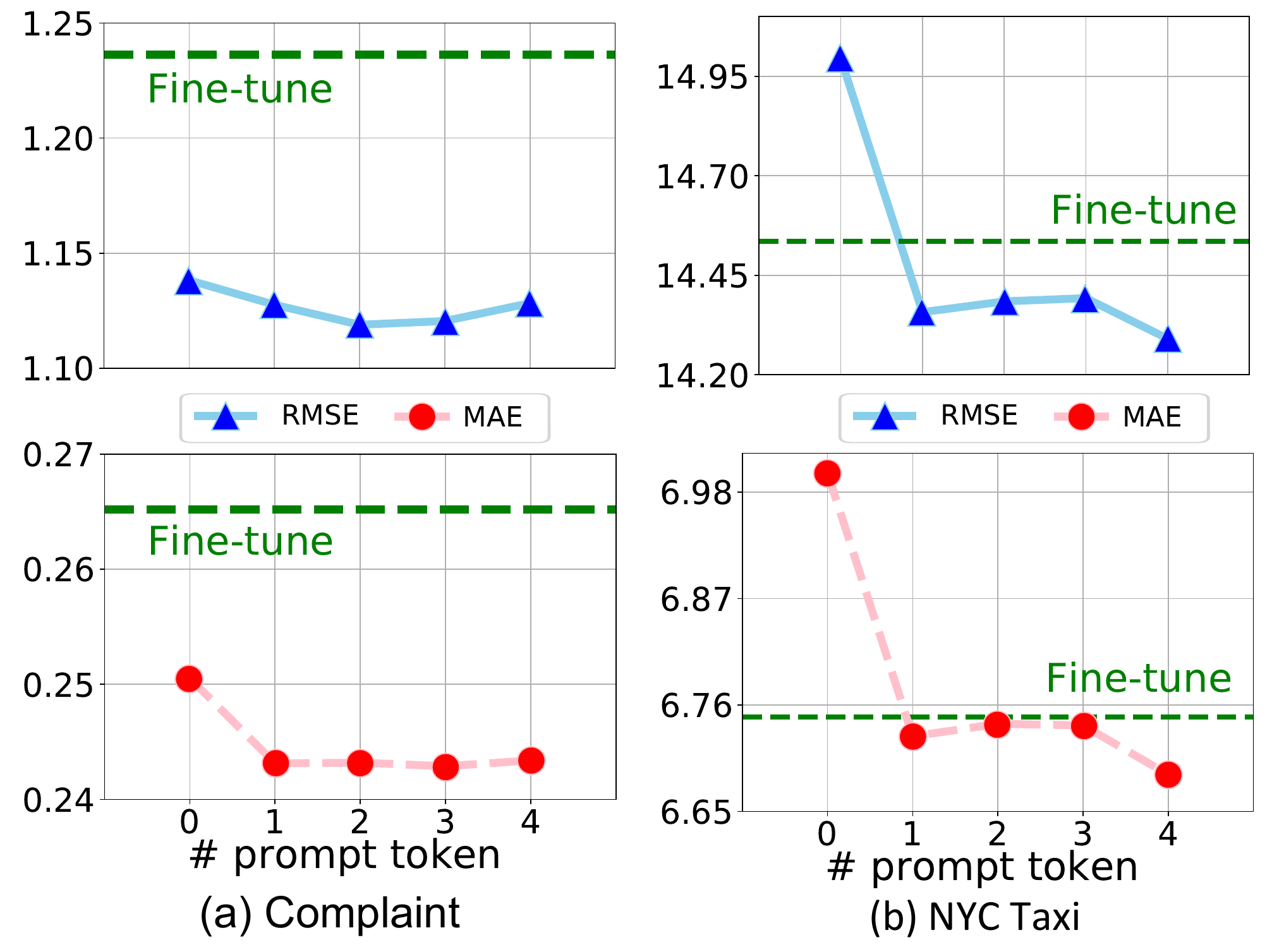}
    \vspace{-5mm}
    \caption{Influence of \stpmt tokens numbers. For a clear comparison, we show the performance of fine-tuning with dashed lines.
    % We present the average RMSE and MAE of all the 19 attributes of Complaint, and the average value of 4 attributes of NYC Taxi.
    }
    \label{fig:hyper}
    \vspace{-5mm}
\end{figure}

% \begin{figure*}[!t]
% {\subfigure{\includegraphics[width=0.16\linewidth]{Figure/pmt0.pdf}}}
% {\subfigure{\includegraphics[width=0.16\linewidth]{Figure/pmt4.pdf}}}
% {\subfigure{\includegraphics[width=0.16\linewidth]{Figure/pmt7.pdf}}}
% {\subfigure{\includegraphics[width=0.16\linewidth]{Figure/pmt11.pdf}}}
% {\subfigure{\includegraphics[width=0.16\linewidth]{Figure/pmt15.pdf}}}
% \vspace{-4mm}

% {\subfigure{\includegraphics[width=0.16\linewidth]{Figure/pmt0_type0.pdf}}}
% {\subfigure{\includegraphics[width=0.16\linewidth]{Figure/pmt4_type0.pdf}}}
% {\subfigure{\includegraphics[width=0.16\linewidth]{Figure/pmt7_type0.pdf}}}
% {\subfigure{\includegraphics[width=0.16\linewidth]{Figure/pmt11_type0.pdf}}}
% {\subfigure{\includegraphics[width=0.16\linewidth]{Figure/pmt15_type0.pdf}}}
%     % \vspace{-3mm}
%     \caption{Figures (a), (b), (c), (d), and (e) present five attributes of Complaint. We compute the total value of each grid and show in an $N-$grids heat map, the darker the color, the larger volume the grid owns. The corresponding prompt tokens are illustrated below as (f), (g), (h), (i), and (j). We calculate the average value of embedding of each grid and darker color means a larger value at the same position in prompt token.
%     }
%     \label{fig:visual}
%     % \vspace{-3mm}
% \end{figure*}

\subsection{Hyper-Parameter Analysis}
To further verify the efficacy of \name, we illustrate how the number of prompt tokens influences the performance of \name in this subsection.
Figure \ref{fig:hyper} indicates the performance of \name with different numbers of \stpmt tokens on the two datasets.
We present the average RMSE and MAE of all the 19 attributes of Complaint, and the average value of 4 attributes of NYC Taxi.
We test \name with $n_{st}$ in $\{0, 1, 2, 3, 4\}$, where $n_{st}=0$ represents no prompt tokens inserted and only tuning head $\boldsymbol{\omega}$.
For a clear understanding, we anchor the result of fine-tune as the red dashed line. 
Firstly, prompt tuning with all $n_{st}$ settings exceeds fine-tune, which proves the steady performance of \name.
Furthermore, different numbers of $n_{st}$ lead to similar results, and we can observe a tiny turn as $n_{st}$ rises.
Take Complaint for an example, \name achieves the best results at $n_{st}=2$, and decreases as $n_{st}$ rises.
The prompt tuning may overfit the target attribute to some extent as the number of prompt tokens rises.

% \subsection{Visualization}
% In \name, we integrate prompt tokens to fit the target spatio-temporal attribute, while fixing the main body of pretrained benchmark.
% In this section, we try to depict the prompt and provide a straight-forward view in a heat map.
% We first pick five spatio-temporal attributes of Complaint dataset, and compute the total volume of such social attributes, as shown in the first line of Figure \ref{fig:visual}.
% At the corresponding positions below each spatio-temporal attribute illustration, we visualize the \stpmt tokens tuned on the spatio-temporal attribute above.
% Specifically, we take the first prompt token ($n_{st}=2$ tokens in total) of the first temporal encoder layer ($l_t=2$ layers in total) with the shape of $\boldsymbol{\mathbb{R}}^{N\times D}$, and sum up the $D$ dimension to attain a $8\times 8$ figure.

% From the figure, we can see that the prompt tokens manifest the conspicuous region to some extent, which reaches consensus with the illustration of corresponding spatio-temporal attributes. 
% Take Figure \ref{fig:visual} (a) as an example, each grid describes the total complaint about commercial noise in this area, and the darker the color, the greater the complaint volume.
% It can be inferred that the most prosperous commercial area locates at the darkest grid.
% Our prompt token in Figure \ref{fig:visual} (f) is acutely aware of this and converges with the highest value at the same position as (a).
% Similar phenomena emerge consistently on other attributes. 

\section{Related Works}

% \subsection{Spatio-Temporal Prediction}
\noindent\textbf{Spatio-Temporal Prediction.}
Spatio-temporal prediction plays a critical role in current city management, such as traffic prediction \cite{yu2017spatio, chai2018bike, wang2021traffic},  weather prediction \cite{han2021joint, zheng2015forecasting, liang2022airformer}, and next-POI recommendation \cite{zhang2021interactive, zhao2016stellar}.
Thanks to the apace boosting of deep learning techniques, recurrent neural networks \cite{ma2015long, tian2015predicting} and 1-D convolutional network \cite{guo2019attention, yu2017spatio} have been widely used in temporal information capture.
Convolutional network \cite{strn} and graph neural network \cite{hgcn,astgcn} are generally incorporated to learn the spatial relationship. 
% Given the multiple fields to be handled in city management, it is necessary to address spatio-temporal multi-attribute prediction.
% However, most current works focus on pursuing performance on single tasks and ignore the close relationship among different spatio-temporal attributes.

% The attention mechanism has proved its efficacy in spatio-temporal prediction recently \cite{traffictmr, liang2022airformer, tmr_icml22, tmr_tkde22, tmr_tnnls22}.
% Traffic Transformer \cite{traffictmr} proposes an encoder-decoder transformer, and incorporates the multi-hop adjacency to mask the feature in transformer decoder.
% DSTAGNN \cite{tmr_icml22} puts forward an adjacency matrix-weighted attention mechanism, and combines it with graph convolutional networks.
% Compared with these entangled architecture, we propose a straight-forward and compact framework consisting of temporal encoder, spatial encoder, and head.
% Our spatio-temporal transformer can address the spatio-temporal characteristic effectively and efficiently. Equipped with our parameter-sharing strategy, it is able to maintain common knowledge among multiple attributes.

\noindent\textbf{Spatio-Temporal Multi-Task Learning.}
% \subsection{Spatio-Temporal Multi-Task Learning}
Recently, attention has been paid to spatio-temporal multi-task learning \cite{zhang2019flow, wang2019origin, zhang2020taxi}.
As one of the earliest efforts, 
GEML \cite{wang2019origin} devises LSTM to predict traffic on-demand flow. 
MDL \cite{zhang2020taxi} employs a convolutional neural network to predict both traffic flow and traffic on-demand flow simultaneously.
Zhang \etal \cite{zhang2019flow} proposes to model the traffic pick-up and drop-off flows respectively with LSTM-based architecture and incorporates an MLP fusion layer for prediction.
What's more, there emerge multi-task learning methods for other spatio-temporal tasks.
MasterGNN \cite{han2021joint} utilizes a heterogeneous recurrent GNN to capture the spatio-temporal correlation between air quality and weather monitor stations.
MATURE \cite{li2020knowledge} and KA2M2 \cite{li2021multi} propose to predict transportation mode with multi-task methods.
% This line of works lie out of the scope of spatio-temporal multi-attribute prediction.

% The existing spatio-tmeporal multi-task learning methods concentrate on solving two specific tasks, and devise a highly-specific framework to address the correlation between the two tasks. 
% Our \name is totally unrestricted to the number of tasks, which is verified on the two real-world datasets, 
% which is the earliest endeavor to address spatio-temporal prediction of numerous attributes.
% Furthermore, it can be easily extended to new attributes by prompt tuning with lightweight \stpmt tokens.

\section{Conclusion}
Spatio-temporal prediction plays a vital role in city management, but the analysis of diverse spatio-temporal attributes requires extensive expert efforts and economic costs. 
% Current works fail to handle the spatio-temporal multivariate prediction due to the heterogeneous spatio-temporal characteristics of the multiple attributes.  
To alleviate this pain spot and explore the solution towards unified smart city modeling, we propose an efficient method, \name to solve spatio-temporal multi-attribute prediction.
% Explicitly, we put forward a compact spatio-temporal transformer model and train on multiple attributes in parallel to capture the common spatio-temporal knowledge. 
% which employs a temporal encoder and spatial encoder to address the temporal and spatial information of spatio-temporal attributes.
% Through training with multiple attributes in parallel, it is able to capture the common spatio-temporal knowledge.  
% Then, we design fabricated \stpmt tokens to effectively fit the heterogeneous knowledge of each attribute and equip the pretrained transformer with efficient prompt tuning.
\name achieves a good balance between fitting the concrete attribute and maintaining the common knowledge.
It also enjoys good transferability, which provides the potential to be extended to new spatio-temporal tasks.
% In the future, we will explore the application of cross-domain multi-attribute prediction.
% In the future, we will explore other efficient tuning method, such as Adapter~\cite{houlsby2019parameter}, SSF~\cite{lianscaling}, \etc, for spatio-temporal multivariate prediction task.

\begin{acks}
This research was partially supported by APRC - CityU New Research Initiatives (No.9610565, Start-up Grant for New Faculty of City University of Hong Kong), CityU - HKIDS Early Career Research Grant (No.9360163), Hong Kong ITC Innovation and Technology Fund Midstream Research Programme for Universities Project (No.ITS/034/22MS), SIRG - CityU Strategic Interdisciplinary Research Grant (No.7020046, No.7020074), SRG-Fd - CityU Strategic Research Grant (No.7005894), Tencent (CCF-Tencent Open Fund, Tencent Rhino-Bird Focused Research Fund), Huawei (Huawei Innovation Research Program), Ant Group (CCF-Ant Research Fund, Ant Group Research Fund) and Kuaishou.
Hongwei Zhao is funded by the Provincial Science and Technology Innovation Special Fund Project of Jilin Province, grant number 20190302026GX, Natural Science Foundation of Jilin Province, grant number 20200201037JC, and the Fundamental Research Funds for the Central Universities, JLU.
Zitao Liu is supported by National Key R\&D Program of China, under Grant No. 2022YFC3303600, and Key Laboratory of Smart Education of Guangdong Higher Education Institutes, Jinan University (2022LSYS003).
\end{acks}

\bibliographystyle{ACM-Reference-Format}
\bibliography{9Reference}

% \clearpage
% \appendix
% \input{8_Appendix.tex}
\end{document}